\documentclass{article}

\usepackage{PRIMEarxiv}

\usepackage[utf8]{inputenc} 
\usepackage[T1]{fontenc}    
\usepackage{hyperref}       
\usepackage{url}            
\usepackage{booktabs}       
\usepackage{amsfonts}       
\usepackage{nicefrac}       
\usepackage{microtype}      
\usepackage{lipsum}
\usepackage{fancyhdr}       
\usepackage{graphicx}       
\graphicspath{{figures/}}     
\usepackage{cite}
\usepackage{ulem}
\usepackage{amsmath}
\usepackage{afterpage}
\usepackage{caption}

\title{Swarms of Large Language Model Agents for Protein Sequence Design with Experimental Validation

}

\author{
  Fiona Y. Wang\textsuperscript{1} \quad
  Di Sheng Lee\textsuperscript{2} \quad
  David L. Kaplan\textsuperscript{2} \quad
  Markus J. Buehler\textsuperscript{3,\#} \\
  \\
  \textsuperscript{1}Laboratory for Atomistic and Molecular Mechanics (LAMM), Department of Biological Engineering, \\
  Massachusetts Institute of Technology, Cambridge, MA 02139, USA \\
  \textsuperscript{2}Department of Biomedical Engineering, Tufts University, Medford, MA, 02155 \\
  \textsuperscript{3}Laboratory for Atomistic and Molecular Mechanics (LAMM), Department of Civil and Environmental Engineering, \\
  Department of Mechanical Engineering, Center for Computational Science and Engineering, \\
  Schwarzman College of Computing, Massachusetts Institute of Technology, Cambridge, MA 02139, USA \\
  \\
  Corresponding author: 
  \textsuperscript{\#}
  \texttt{mbuehler@mit.edu}
}

\begin{document}
\maketitle
\begin{abstract}
Designing proteins \textit{de novo} with tailored structural, physicochemical, and 
functional properties remains a grand challenge in biotechnology, medicine, and materials 
science, due to the vastness of sequence space and the complex coupling between sequence, 
structure, and function. Current state-of-the-art generative methods, such as protein 
language models (PLMs) and diffusion-based architectures, often require extensive 
fine-tuning, task-specific data, or model reconfiguration to support objective-directed 
design, thereby limiting their flexibility and scalability. To overcome these limitations, 
we present a decentralized, agent-based framework inspired by swarm intelligence for 
\textit{de novo} protein design. In this approach, multiple large language model (LLM) 
agents operate in parallel, each assigned to a specific residue position. These agents 
iteratively propose context-aware mutations by integrating design objectives, local 
neighborhood interactions, and memory and feedback from previous iterations. This 
position-wise, decentralized coordination enables emergent design of diverse, 
well-defined sequences without reliance on motif scaffolds or multiple sequence alignments, 
validated with experiments on proteins with alpha helix and coil structures. 
Through analyses of residue conservation, structure-based metrics, and sequence convergence 
and embeddings, we demonstrate that the framework exhibits emergent behaviors and effective 
navigation of the protein fitness landscape. Our method achieves efficient, objective-directed 
designs within a few GPU-hours and operates entirely without fine-tuning or specialized training, 
offering a generalizable and adaptable solution for protein design. Beyond proteins, the approach 
lays the groundwork for collective LLM-driven design across biomolecular systems and other scientific discovery tasks.
\end{abstract}

\keywords{protein design \and swarm intelligence \and large language models \and agent-based systems}

\section{Introduction}
The ability to computationally design novel proteins with precisely tailored structural, 
physicochemical, and functional properties is a grand challenge at the forefront of modern 
biotechnology\cite{RN5, RN6, RN7, RN8, RN9, RN14}, holding immense promise for advancements 
across medicine\cite{RN10, RN11, RN13}, materials science\cite{RN15, RN17, RN18}, and 
synthetic biology\cite{RN19, RN20}. \textit{De novo} protein design, which aims to create 
entirely new amino acid sequences that fold into desired three-dimensional structures and 
perform specific tasks, offers a powerful alternative to modifying existing natural 
proteins\cite{RN21}. However, navigating the astronomically vast protein sequence space, 
coupled with the intricate and often non-intuitive relationship between sequence, structure, 
and function, renders this a formidable computational and experimental endeavor.

Traditional computational protein design methods have made significant strides, often relying on 
physics-based energy functions or statistical potentials to guide sequence optimization within a 
predefined structural scaffold\cite{RN22, RN23}. More recently, the advent of deep learning has 
revolutionized protein science, giving rise to powerful neural networks\cite{RN25, RN24, RN29, RN2, RN57}, 
protein language models (PLMs)\cite{RN3, RN26, RN27, RN28}, and denoising diffusion probabilistic 
models\cite{RN1, trippe2022diffusion, diffusion}. While these models excel at tasks like protein 
folding prediction or generating natural-like sequences, their application to targeted 
\textit{de novo} protein design often presents significant limitations. Specifically, many 
state-of-the-art generative models typically demand extensive fine-tuning on large, task-specific 
datasets or intricate architectural modifications to achieve multi-objective design (Figure ~\ref{fig:overview}).
This reliance 
on specialized training data and computationally intensive learning phases restricts their 
generalizability, adaptability to novel design objectives, and overall efficiency for rapid prototyping 
in diverse design scenarios. Consequently, a significant gap remains in developing highly adaptable, 
generalizable, and computationally efficient \textit{de novo} protein design frameworks that do not 
require extensive pre-training or fine-tuning for each new objective.

Large Language Models (LLMs), powerful neural networks initially designed for natural language processing tasks, have demonstrated an unprecedented ability to learn complex patterns, capture intricate relationships within sequential data, and generate coherent, contextually relevant outputs\cite{NIPS2017_3f5ee243, NEURIPS2020_1457c0d6}. Their success stems from their transformer-based architectures, which enable them to process long-range dependencies and learn rich representations from vast amounts of unsupervised data. This inherent capacity for pattern recognition and sequence generation has naturally extended their application to diverse scientific domains, including the automation of scientific discovery through bio-inspired multi-agent intelligent graph reasoning\cite{RN30}, the exploratory optimization of reasoning and agentic thinking\cite{RN31}, the discovery of protein design principles\cite{ghafarollahi2025sparksmultiagentartificialintelligence}, protein discovery via physics- and machine learning-informed multi-agent collaborations\cite{D4DD00013G}, the solving of complex mechanics problems and knowledge integration\cite{RN32}, and molecular analysis and design using generative AI via multi-agent modeling\cite{D4ME00174E}. 

To address limitations of the current computational protein design methods and leveraging the advantages of LLMs, we introduce a novel, decentralized, and collaborative agent-based framework for \textit{de novo} protein sequence design, drawing inspiration from the principles of swarm intelligence\cite{rahman2025llmpoweredswarmsnewfrontier}. Our approach harnesses the emergent collective behavior of multiple LLM agents, each assigned to a specific residue position within the protein sequence. These LLM agents iteratively propose context-aware mutations, integrating rich information including explicit design objectives, fundamental protein folding principles, local neighborhood information, a memory of prior iterations, and evaluation to guide their decisions. This unique position-wise, decentralized strategy facilitates the emergent design of structured, diverse, and multi-objective protein sequences without the need for explicit motif scaffolds or resource-intensive multiple sequence alignments.

\section{Results and Discussion}
\subsection{Swarm Framework}
\begin{figure}[htbp]
    \centering
    \includegraphics[width=\textwidth]{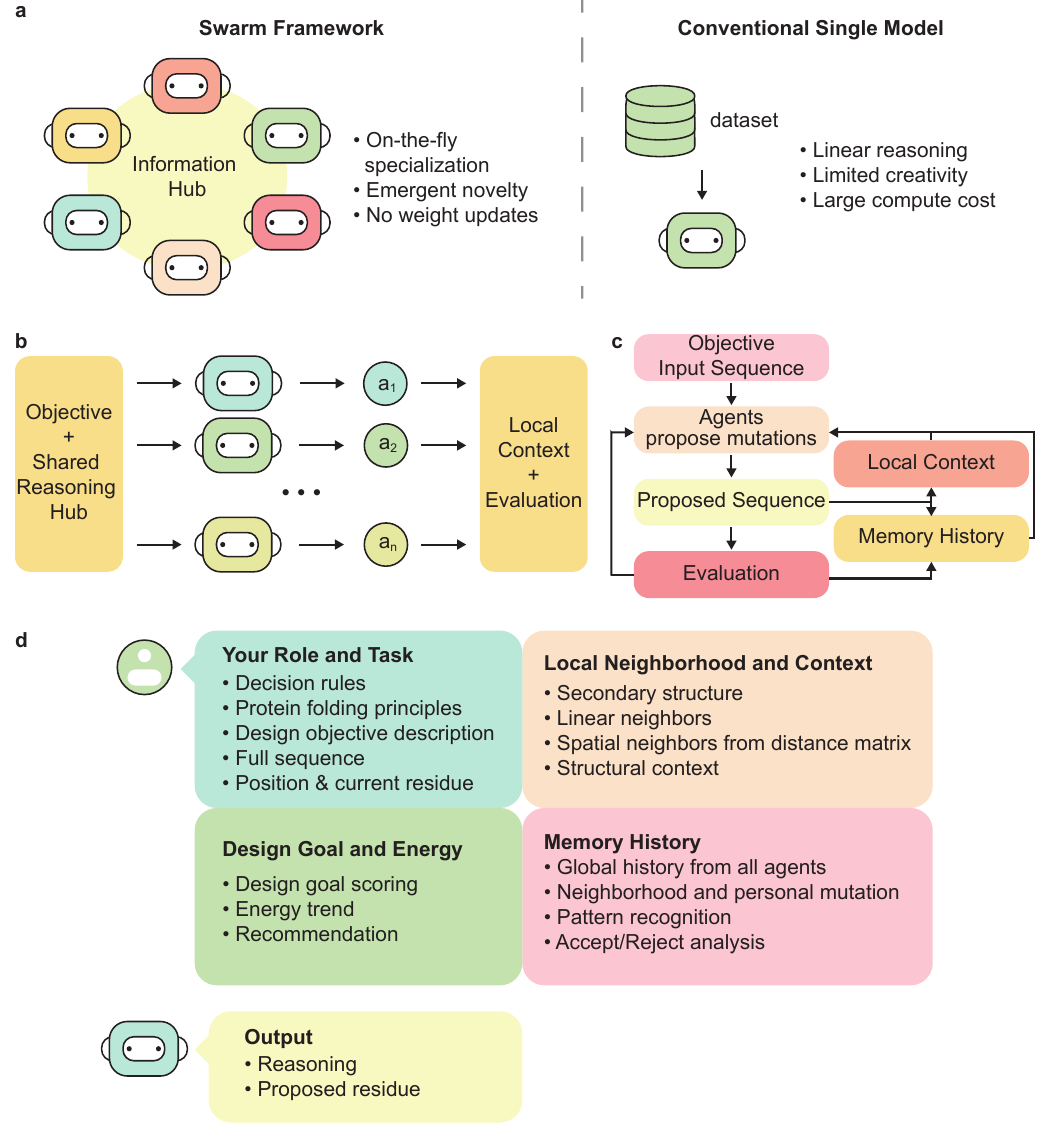}
    \caption{\textbf{a}. Comparison of swarm framework with conventional
    single-model framework. \textbf{b}. Multiple LLM agents share the same design objective and 
    reasoning hubs, each proposing mutations for a single residue position, producing updated local context
    and evaluation feedback. \textbf{c}. Starting with a design objective and input sequence, 
    the agents propose mutations for each residue position, producing an updated sequence which is
    evaluated. The previously proposed sequences are stored in memory for future iterations. The local context, 
    memory history, and evaluation feedback are used to guide the next round of mutations. 
    \textbf{d}. Input prompt consists of the agent's role and task, local neighborhood and context, 
    design goal and energy, and memory history. Output consists of reasoning and the proposed mutation.
    }
    \label{fig:overview}
\end{figure}

Our protein sequence design framework leverages a decentralized, agent-based approach inspired 
by swarm intelligence (swarm framework), where individual LLM agents collaboratively optimize a protein sequence 
to meet predefined design objectives. The overall framework, depicted in Figure~\ref{fig:overview}b,
comprises of an objective and shared reasoning hubs, a group of parallel LLM agents, 
and an evaluation module.

The process begins with the user-defined objective and input sequence (Figure~\ref{fig:overview}c). 
This information is passed to the group of parallel LLM agents (Figure~\ref{fig:overview}b), each 
in charge of a single residue position, e.g. $a_1, a_2, \ldots, a_n$. Rather than relying on costly weight updates, these agents perform on-the-fly specialization, using 
the information provided by the objective, shared reasoning hubs, local context, and evaluation feedback to propose mutations for the current sequence (Figures S1-S5).
By concatenating the proposed mutations from all agents, we obtain an updated sequence, which is then
passed to the evaluation module, which converts the sequence into a PDB file and evaluates it against the design objectives.
Local context is extracted from the PDB file and passed to the next iteration, allowing the swarm to learn from its history and generate emergent novelty.
Memory history is extracted from the proposed sequence and its corresponding evaluation results, allowing the swarm to learn from its past decisions.
The proposed sequence, local context, memory history, and evaluation feedback are then fed back into the prompt for the next iteration,
allowing the LLM agents to collectively explore the design space, learn from its history and neighborhood, and improve on the evaluation results to achieve the final optimized sequence.

\subsection{Swarm Framework Achieves Diverse Structural Objectives}

\begin{figure}[htbp]
    \centering
    \includegraphics[width=0.85\textwidth]{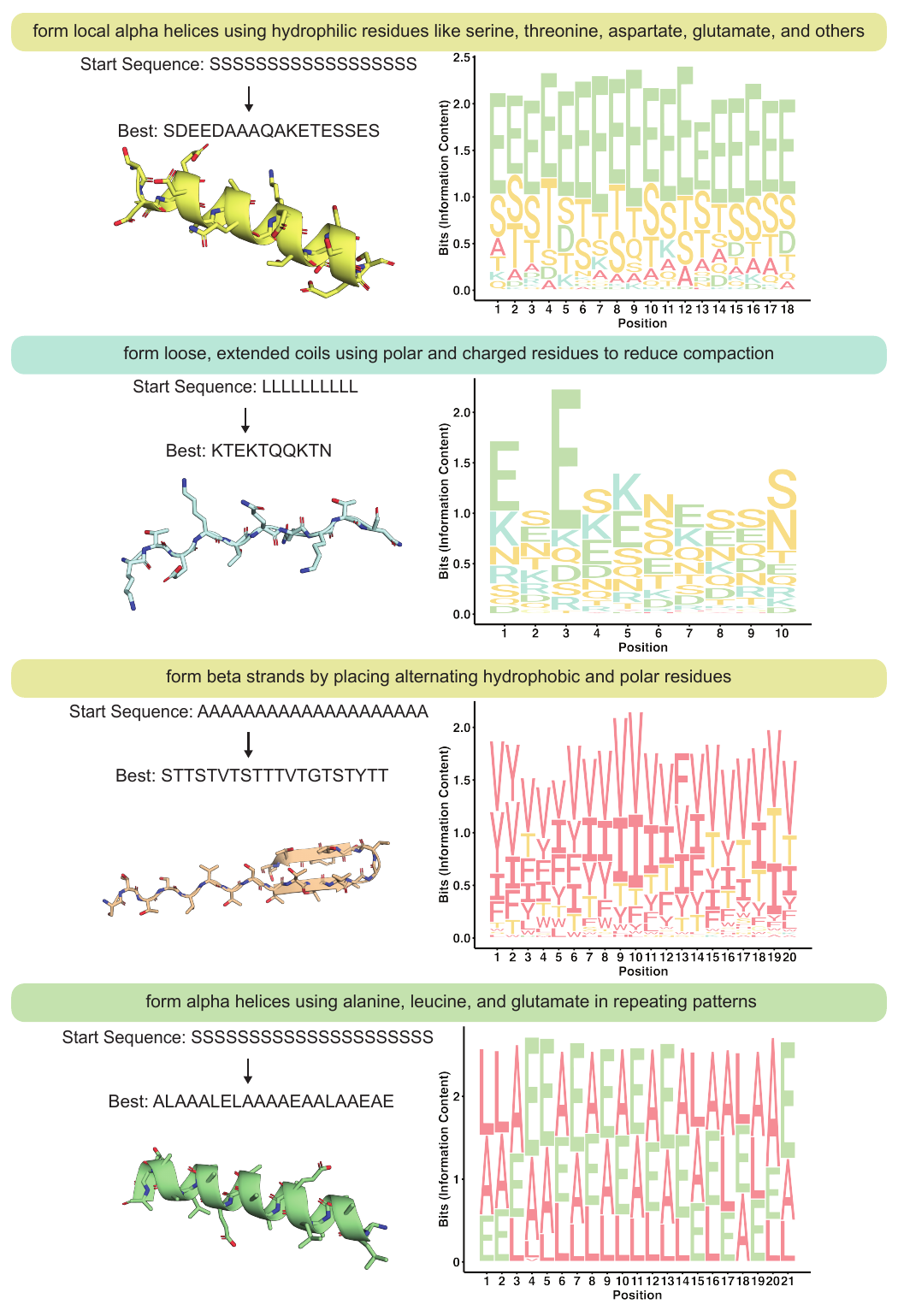}
    \caption{Design objective, start sequence, best sequence, its respective 3D structure, and sequence logo returned from 64 iterations with GPT-4o for four structural design objectives.}    
    \label{fig:best_sequences}
\end{figure}

As shown in Figure~\ref{fig:best_sequences}, the proposed framework generates 
protein sequences that demonstrate diverse secondary structures in accordance 
with predefined design objectives.

When challenged to form $\alpha$-helices from an initial SSSS... sequence, 
the framework generates helical structures under two constraints: one set 
containing a diversity of hydrophilic residues (top panel) and another with 
repeating alanine, leucine, and glutamate (ALE) motifs (bottom panel). 
The corresponding sequences (SDEEDAAAQAKETESSES and ALAAALELAAAAEAALAAEAE) and sequence logos illustrate these preferences, 
showing enrichment for hydrophilic helical residues for the top panel,
and enrichment for the ALE pattern for the bottom panel, 
in alignment with well-known helix stabilization rules\cite{RN35, RN36}. The framework also successfully engineers other topologies, 
as seen by its ability to design a sequence (STTSTVTSTTTTVIGTSTYYT) 
that forms a beta-strand by placing alternating hydrophobic and polar residues. 
The sequence logo illustrates the enrichment for hydrophobic (red) and polar (yellow) residues, 
in alignment with principles of $\beta$-sheet formation\cite{RN38}.
Furthermore, it can optimize for random structures, such as generating 
a "loose, extended coil" (KTEKTQQKTN) from a hydrophobic LLLL... sequence. 
The sequence logo illustrates the enrichment for charged (green and blue) and polar (yellow) residues, 
in alignment with principles of coil formation\cite{RN42, RN43}.
In each example, the 3D-folded structures confirm that the sequence indeed
achieves the desired structural motif.

To experimentally validate that the designed sequences indeed form the 
intended secondary structure motifs, circular dichroism (CD) 
spectroscopy was performed on two hydrophilic peptide examples that can be readily synthesized: 
the hydrophilic $\alpha$-helix and the coil sequence. 
CD spectroscopy is a standard biophysical technique for probing the secondary 
structure content of peptides and proteins in solution\cite{RN58}.

\begin{figure}[htbp]
    \centering
    \includegraphics[width=\textwidth]{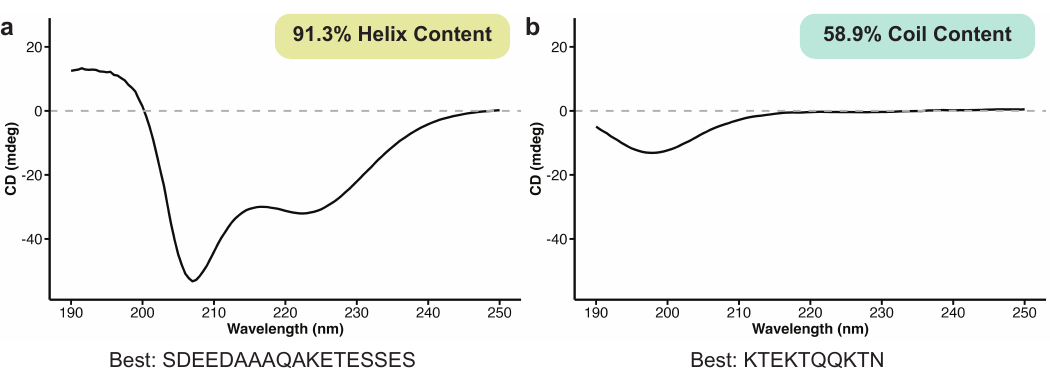}
    \caption{CD spectra of the best sequence for the \textbf{a}. hydrophilic helix and the \textbf{b}. coil sequence. }
    \label{fig:CD}
\end{figure}

For the hydrophilic helix design, the CD spectrum in Figure~\ref{fig:CD}a displayed characteristic 
double minima near 208~nm and 222~nm, which is a hallmark of $\alpha$-helical 
conformation\cite{RN59}. This confirms that the designed sequence adopts 
a helical structure in aqueous solution, in agreement with the in silico prediction. Similarly, the sequence designed to form a random coil exhibited in Figure~\ref{fig:CD}b
a very low ellipticity above 210~nm and a negative band near 195~nm, 
consistent with the spectral signature of a predominantly disordered 
(random coil) conformation\cite{RN60}. These experimental results validate that the swarm framework can 
produce sequences with robust secondary structure content. 

 \begin{figure}[htbp]
    \centering
    \includegraphics[width=0.7\textwidth]{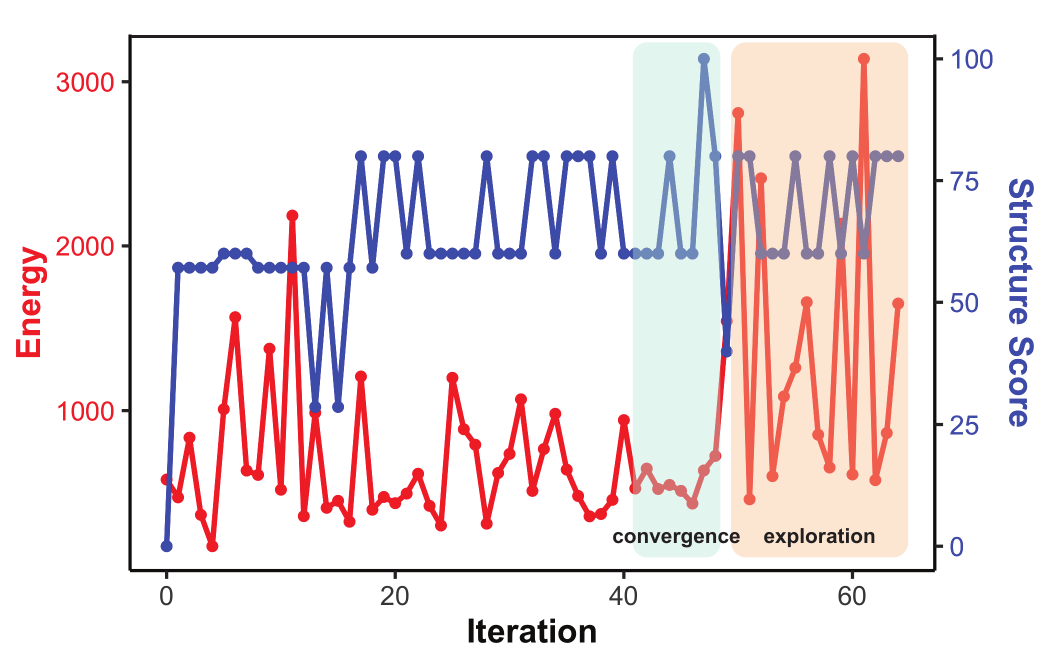}
    \caption{Evolution of calculated Rosetta energy\cite{RN34} (red) and 
    Structure Score (blue) over 64 iterations with GPT-4o for the design objective: 
    choose residues that mirror their left and right neighbors to promote local symmetry. 
    This plot visualizes the dynamic interplay between convergence (blue shaded 
    regions, where Structure Score stabilizes at a high level and Energy 
    stabilizes at a low level) and exploration (orange shaded regions, 
    where Energy fluctuates more significantly) during the iterative design process.}
    \label{fig:convergence}
\end{figure}

Interestingly, the iterative design process, powered by LLMs such as GPT-4o\cite{openai_chatgpt}, reveals a dynamic balance between convergence and exploration over 64 iterations, as illustrated in Figure~\ref{fig:convergence}. This figure tracks Rosetta Energy\cite{RN34, RN44, RN45} (red), which reports the physical plausibility of each structure, and Structural Score (blue), which reflects how closely the predicted fold matches the target motif. Early iterations (blue, 40-45) push the sequence rapidly toward high Structural Scores (often >75\%) while lowering Rosetta Energy; later iterations (orange, 48-64) keep the motif intact yet continue probing alternative, lower-energy sequences, underscoring the framework's alternating converge-and-explore strategy\cite{ma2025explorllmguidingexplorationreinforcement}.

\subsection{Swarm Framework Achieves Physiochemical, Functional, and Multi-Domain Objectives}
Beyond structural motifs, the swarm framework exhibits 
strong adaptability in achieving physiochemical, functional, and multi-domain 
design objectives such as matching specific vibrational frequency distributions, 
designing proteins with metal-binding pockets, and designing proteins 
with multi-domain structures.
\begin{figure}[htbp]
    \centering
    \includegraphics[width=\textwidth]{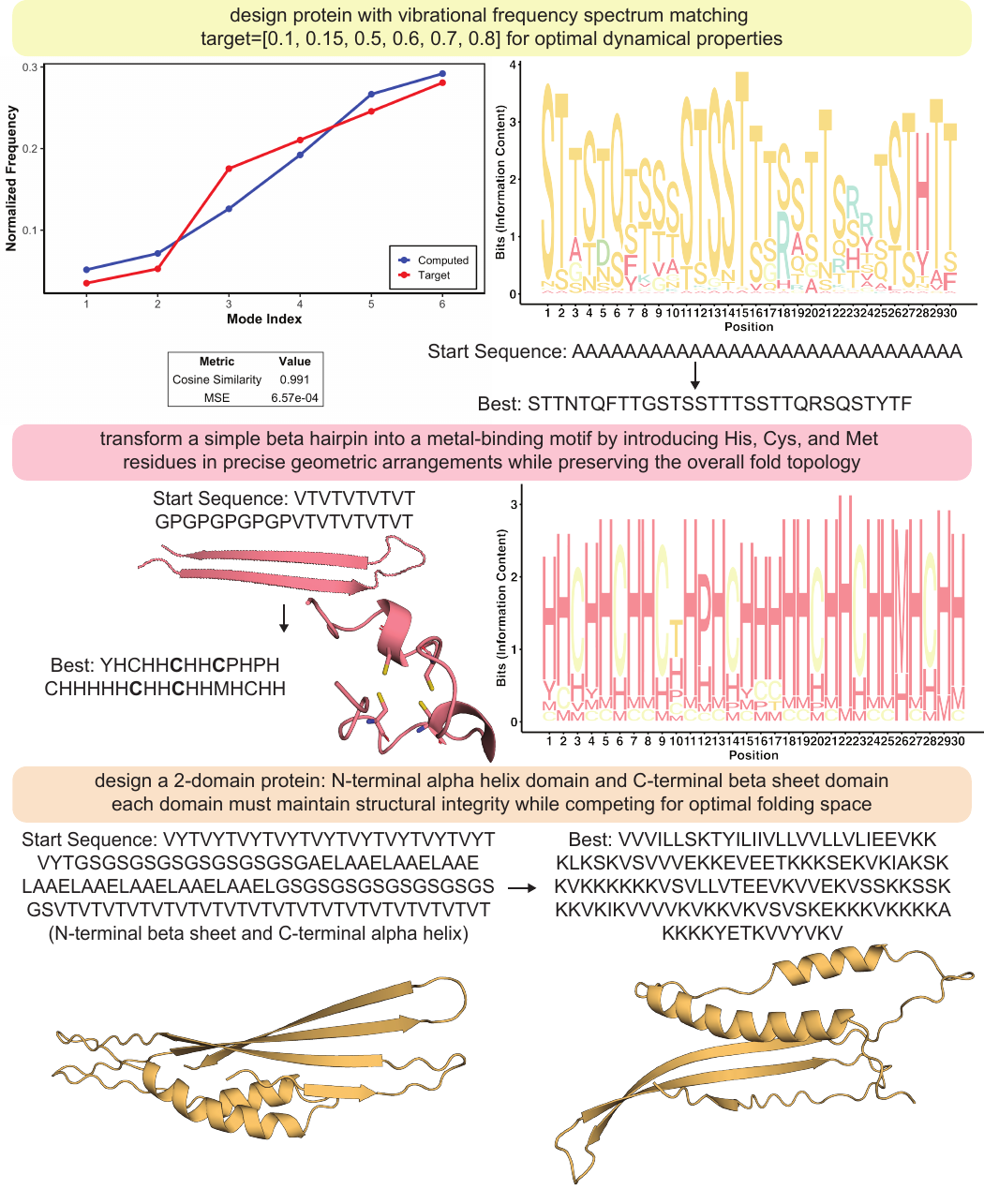}
    \caption{Design objective, start sequence, best sequence, evidence of objective achievement, and sequence logo returned from 16 iterations with GPT-4o for three diverse design objectives.}    
    \label{fig:more_design_objectives}
\end{figure}


As shown in the top panel of Figure~\ref{fig:more_design_objectives}, 
the framework accurately reproduces a predefined target normalized frequency distribution 
across six mode indices, which represent the lowest-frequency collective motions 
obtained from an Anisotropic Network Model (ANM) of the protein\cite{RN46}. 
These frequencies correspond to the square roots of the eigenvalues derived from the 
Hessian matrix of the ANM\cite{RN46}. The computed frequencies of the designed protein (blue) closely align with the 
target frequencies (red), with a cosine similarity of 0.991 and a mean squared error (MSE) of 6.57e-04, 
approaching almost perfect agreement, highlighting the precision with which the swarm
framework can optimize for frequency distribution objectives. 

\begin{figure}[htbp]
    \centering
    \includegraphics[width=0.7\textwidth]{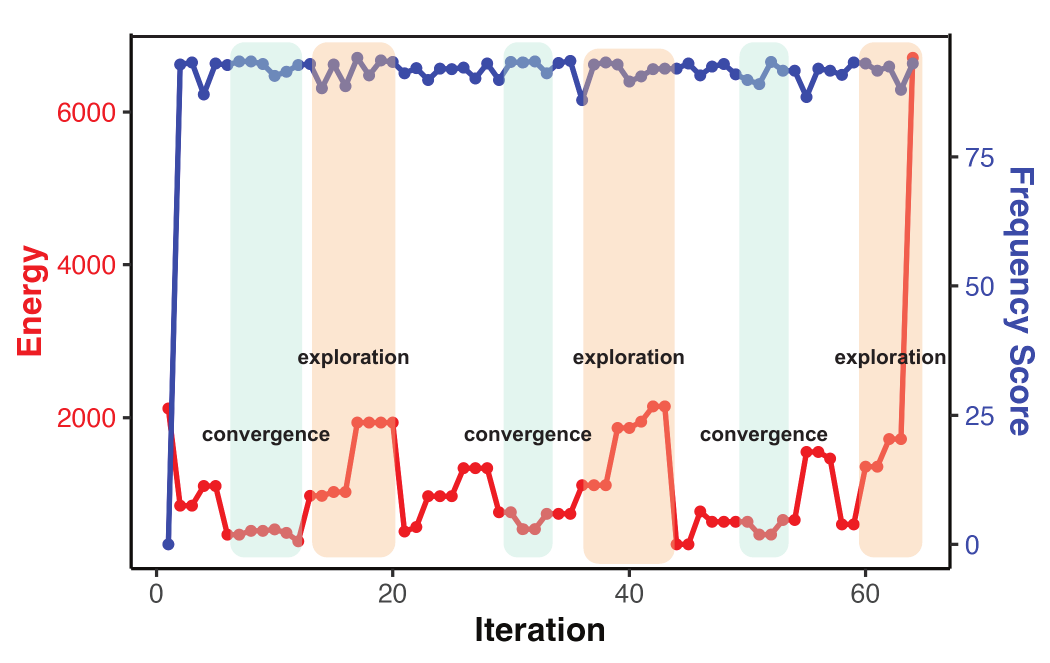}
    \caption{Evolution of calculated Rosetta energy\cite{RN34} (red) and Frequency Score (blue) over 64 iterations with GPT-4o for the design objective: design protein with vibrational frequency spectrum matching target=[0.1, 0.15, 0.5, 0.6, 0.7, 0.8]. This plot illustrates the multi-round convergence (blue shaded regions, where Frequency Score is high and stable and Energy is low) and exploration (orange shaded regions, where Frequency Score and Energy fluctuates) dynamics of the swarm framework over multiple iterative cycles for a non-structural design objective.}
    \label{fig:freq_convergence}
\end{figure}

Further insight into the design process of target frequency distribution is 
provided in Figure~\ref{fig:freq_convergence}, which visualizes the evolution of 
Energy (blue) and Frequency Score (red) over time. Three distinct phases of convergence 
(blue) alternate with exploratory phases (orange), where energy 
fluctuations increase as the agents sample broader regions of sequence space. 
This dynamic oscillation between refinement and exploration enables the framework 
to identify optimized solutions while maintaining diversity
in the designed sequences.


The middle panel of Figure~\ref{fig:more_design_objectives} presents the successful transformation
from a beta hairpin into a metal-binding motif by introducing histidine, cysteine, and methionine residues.
The highlighted four cysteine residues (shown in sticks) form a coordination pocket for the metal ion, 
which is consistent with the design objective. The sequence logo highlighted common metal-binding motifs such as CXXC\cite{RN61} without additional knowledge input, 
highlighting the framework's ability to achieve functional objectives with minimal guidance.

The bottom panel of Figure~\ref{fig:more_design_objectives} presents the transformation
of a protein with N-terminal beta sheet and C-terminal alpha helix into a protein with N-terminal 
alpha helix and C-terminal beta sheet, which is consistent with the design objective. 
This indicates the framework is able to design longer sequences (136 residues) and potentially longer
sequences when memory space permits.

\subsection{LLM Comparison}
\begin{figure}[htbp]
    \centering
    \includegraphics[width=\textwidth]{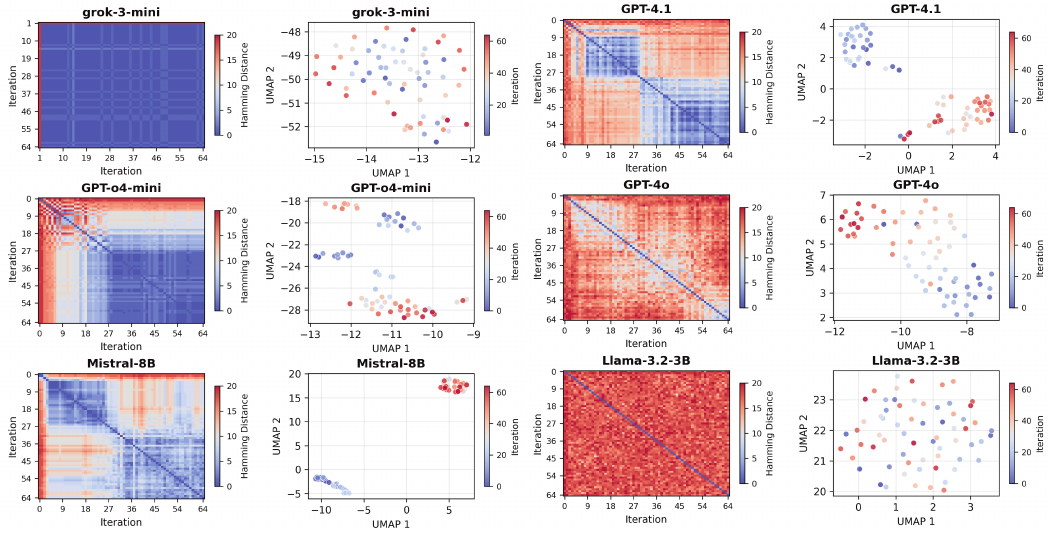}
    \caption{Comparison of 6 LLMs for the local symmetry design objective. The left panels show Hamming distance heatmaps of sequences from 64 iterations, illustrating mutation convergence and diversification (darker colors indicate higher convergence, lighter (red) colors indicate less convergence). The right panels show UMAP clustering of  sequences from 64 iterations based on physicochemical properties, with points colored by iteration, visualizing the exploration and convergence patterns for grok-3-mini, GPT-o4-mini, Mistral-8B, GPT-4.1, GPT-4o, and Llama-3.2-3B.}
    \label{fig:model_comparison}
\end{figure}

The choice of the underlying LLM substantially affects both convergence dynamics and exploration diversity within the swarm framework, as demonstrated in Figure~\ref{fig:model_comparison}. This figure provides a comparative evaluation 
across six distinct language models: grok-3-mini, GPT-4o-mini, Mistral-8B, GPT-4.1, GPT-4o, and Llama-3.2-3B, under the same local symmetry design objective. 
For each model, two visualizations are presented: on the left, a Hamming distance heatmap quantifies sequence similarity 
and convergence, while on the right, a UMAP projection clusters sequences based on their physicochemical properties, across design iterations.

The Hamming distance heatmaps (left panels of Figure~\ref{fig:model_comparison}) visualize pairwise dissimilarity between sequences across iterations. Darker regions indicate smaller Hamming distances (higher similarity), while lighter regions represent greater diversity. Examining these plots reveals that different language models lead to distinct convergence profiles in the sequence space, influenced by each model’s reasoning style, architectural traits, and training distributions:
\begin{itemize}
    \item[\textbf{grok-3-mini}] This model exhibits the strongest convergence behavior, 
    with its heatmap showing uniformly dark shades, 
    indicating that sequence diversity rapidly collapses. 
    This is likely due to both its relatively small model parameter 
    size and the low setting of \texttt{reasoning\_effort = low}, 
    which together result in shallow sampling, limited hypothesis testing, 
    and reduced capacity to explore multiple local optima\cite{grok3mini2025}.
    \item[\textbf{GPT-o4-mini}] Although also a small model, 
    GPT-o4-mini\cite{gpto4mini, gpto4mini-intro} demonstrates slightly more exploration 
    early on before converging. Its pattern reflects a stronger prior for
    structured reasoning than grok-3-mini, enabling initial variation followed 
    by deterministic settling. 
    This balance may stem from more optimized architectural tuning in the GPT-o4-mini compared to grok.
    \item[\textbf{Mistral-8B}] Mistral displays structured convergence 
    with well-defined similarity blocks, indicating consistent and deterministic behavior. 
    Its early-phase trajectories are less diverse, leaning towards convergence 
    than exploration, consistent with its robust pattern-matching capabilities\cite{mistral_models_2024, mistral8B}.
    \item[\textbf{GPT-4.1}] GPT-4.1 exhibits less convergence 
    than Mistral-8B, with clearer signs of early-stage exploration and 
    multiple regions of moderate similarity, 
    suggesting a more distributed search strategy. 
    Its abstraction and generalization capabilities may enable it to explore multiple promising directions in parallel before converging\cite{gpt41}.
    \item[\textbf{GPT-4o}] GPT-4o demonstrates more exploratory behavior, 
    as its Hamming distance heatmap shows a gradual gradient of similarity interweaved by intermittent convergent steps. 
    This pattern reflects its real-time reasoning and multimodal 
    flexibility which allows the model to adaptively explore input space before committing to final outputs\cite{GPT4o}. 
    Its ability to directly integrate and process rich sensory data enables a more nuanced 
    exploration-convergence balance compared to models constrained to unimodal pipelines\cite{gpt4o-card}.
    \item[\textbf{Llama-3.2-3B}] Llama-3.2 shows the highest level of 
    exploration, with diffuse and lightly saturated patterns indicating 
    minimal convergence. This sustained diversity may arise from weaker 
    inductive biases or alignment constraints, favoring broader sampling at the cost 
    of slower optimization toward design objectives\cite{grattafiori2024llama3herdmodels, meta-llama_Llama3.2‑3B_2024, song2025begunhalfdonelowresource}.
\end{itemize}

The UMAP clustering plots (right panels of Figure~\ref{fig:model_comparison}) offer a dimensionality-reduced representation of the generated sequences, projected according to their physicochemical properties. Points are color-coded by iteration, enabling a comparative view of how each language model navigates, samples, and ultimately converges within the sequence landscape over time.

\begin{itemize}
    \item[\textbf{grok-3-mini}] The UMAP projection reveals a single, loosely defined 
    cluster, with minimal differentiation between early and late iterations. This high 
    degree of convergence in physicochemical space likely reflects the model’s relatively 
    low reasoning effort, resulting in a more deterministic and less exploratory trajectory 
    toward a stable solution.
    \item[\textbf{GPT-4o-mini}] Convergence occurs around iteration 27, after which 
    sequences segregate into two distinct clusters. This pattern suggests an initial 
    exploratory phase followed by bifurcation into two dominant
    local minima within the physicochemical space.
    \item[\textbf{Mistral-8B}] Two well-defined convergence clusters occur from 
    iterations 2–25 and 28–64. The clear separation between early and late
    iteration clusters suggests that the model identifies two distinct regions 
    in the design space, with an intermediate exploratory phase between them.
    \item[\textbf{GPT-4.1}] Two convergence regions are observed from iterations
    5–27 and 29–64 though the separation between clusters is less distinct 
    than that of other models. This suggests a more gradual
    transition between solution spaces, consistent with a less discrete 
    and more continuous mode of convergence.
    \item[\textbf{GPT-4o}] Three convergence phases are evident from 
    iterations 2–27, 27–52, and 53–64. However, clusters are less sharply defined, 
    suggesting a more continuous convergent process. 
    This fluid progression aligns with GPT-4o’s real-time reasoning and 
    multimodal adaptability\cite{GPT4o}.
    \item[\textbf{Llama-3.2-3B}] This model exhibits minimal convergence, 
    with early and late sequences distributed homogeneously across the UMAP space. 
    The lack of cluster formation reflects a persistent exploratory behavior, 
    supported by the model’s diffuse sampling strategy across the design landscape\cite{grattafiori2024llama3herdmodels}.
\end{itemize}

Our comparative analysis of LLMs using both Hamming distance heatmaps and UMAP 
clustering reveals diverse behaviors in convergence and exploration, 
which influences the trajectory of protein sequence optimization within our 
swarm framework. Our results demonstrate that the balance between 
convergence and exploration can be tuned by selecting the appropriate LLM:
models like grok-3-mini yield highly convergent optimization, while choices such as 
Llama-3.2-3B favor broader, more exploratory search in sequence space. 
Thus, model selection provides a practical lever to modulate search dynamics in protein 
design.

Behavioral differences between models stem not only from architecture but also 
from transparency. Proprietary models like GPT series are “black boxes,”
limiting reproducibility\cite{RN47, Wolfe_2024, lande_strashnoy_2023_semantic}. 
In contrast, open-weight models like Mistral-8B and Llama-3.2-3B provide more transparency, 
but still withhold key details\cite{jiang2023mistral7b, touvron2023llama2openfoundation}. 
Thus, the convergence behavior observed in swarm optimization is shaped by 
both the internal architecture of the model and its transparency\cite{sapkota2025comprehensiveanalysistransparencyaccessibility}.
The precise mechanistic basis underlying these convergence differences can only be determined 
through further investigation, a current limitation of our study.

\subsection{Benchmark}
\begin{figure}[htbp]
    \centering
    \includegraphics[width=\textwidth]{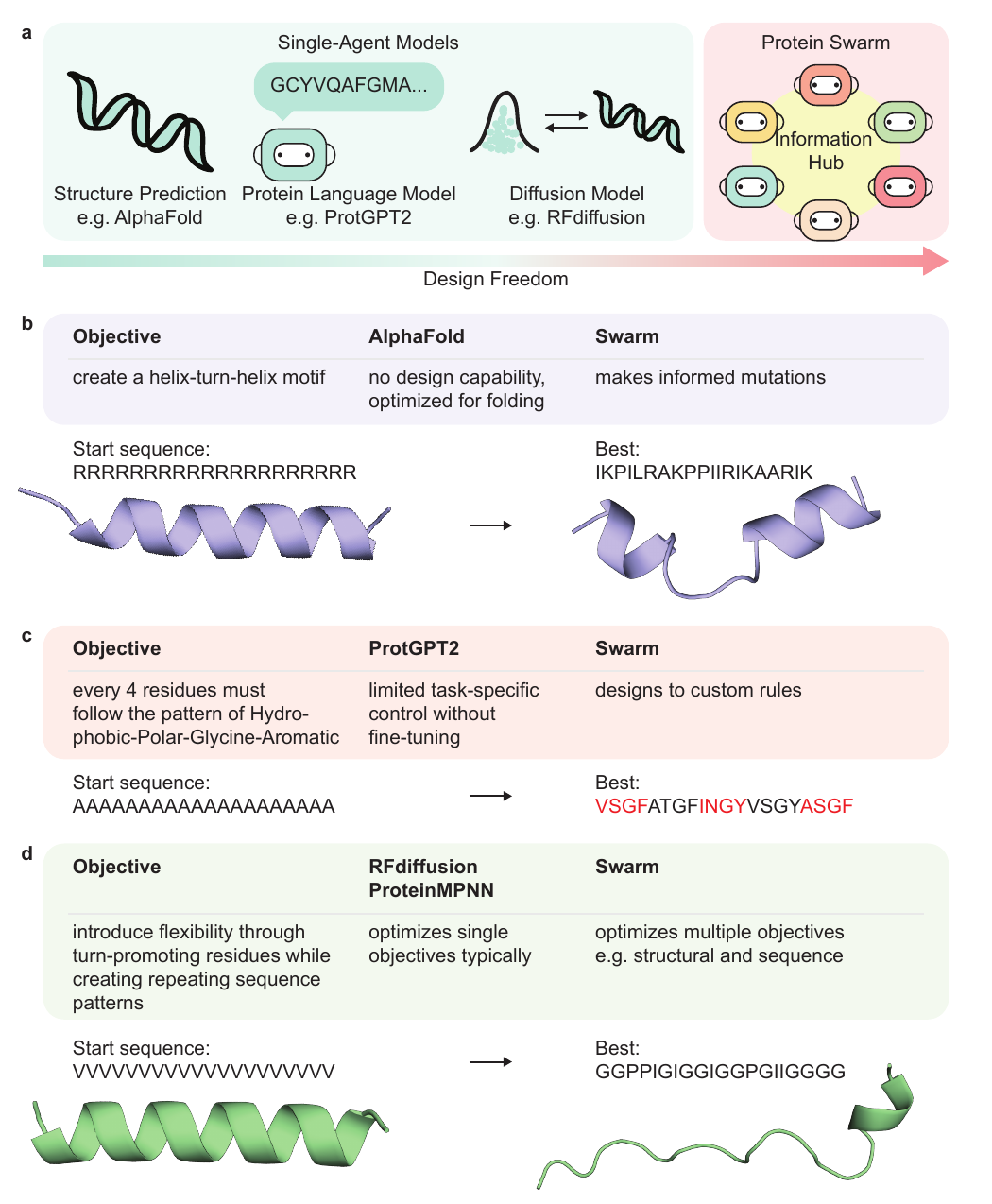}
    \caption{\textbf{a}. Design freedom increases from structural prediction models, autoregressive protein language models, denoising diffusion probabilistic models to swarm framework. 
    \textbf{b}. The swarm framework is capable of making informed mutations compared to static structural prediction models.
    \textbf{c}. The swarm framework can design sequences according to custom rules compared to autoregressive protein language models.
    \textbf{d}. The swarm framework optimize multiple objectives compared to denoising diffusion probabilistic models.}
    \label{fig:generalization_comparison}
\end{figure}

To validate the efficacy and unique advantages of our swarm framework, 
we evalutaed its performance against several established protein design methodologies, 
including structural prediction models, autoregressive protein language models (PLMs), 
and denoising diffusion probabilistic models. 
As shown in Figure~\ref{fig:generalization_comparison}a, 
design freedom increases progressively across the spectrum of protein engineering 
approaches: starting with structural prediction models, 
which allow the least flexibility by evaluating fixed sequences; 
then to PLMs, which can generate sequences but are generally limited to 
patterns learned from natural proteins; 
followed by denoising diffusion probabilistic models, 
which offer more generative capability for sequence and backbone design. Our swarm framework builds upon and exceeds these levels of design freedom, 
providing the ability to impose diverse objectives without 
requiring retraining or modification for new tasks.

Figure~\ref{fig:generalization_comparison}b compares swarm framework’s ability 
to design a helix–turn–helix motif to structural models such as 
AlphaFold\cite{RN48, RN49}. While AlphaFold excels in protein structure prediction, 
it is not inherently suited for \textit{de novo} sequence design. 
Beginning from a poly-arginine sequence, the swarm framework applied 
structure-informed mutations to yield a novel sequence (IKPILRAKPPIIRIKAARIK) that, 
when folded using AlphaFold, adopted the desired helix–turn–helix conformation. 
This result demonstrates the generative power of the swarm framework in 
producing sequences that fold into specific target structures.

Figure~\ref{fig:generalization_comparison}c compares the swarm framework to 
autoregressive PLMs such as ProtGPT2\cite{RN3} in designing sequences with specific 
patterns. While ProtGPT2 generates natural-like sequences, it offers limited 
control over explicit design constraints without substantial fine-tuning. 
In contrast, when assigned a specific sequence design objective of "every 4 residues 
must follow the pattern of Hydrophobic–Polar–Glycine–Aromatic", the swarm framework 
successfully produced a sequence (VSGFATGFINGYVSGYASGF) that strictly adheres to this rule.
This highlights the swarm framework’s capacity to incorporate user-defined, custom rules without 
requiring retraining or architectural changes.

Figure~\ref{fig:generalization_comparison}d compares the swarm framework against 
RFdiffusion\cite{RN1}, a representative denoising diffusion model for 
protein backbone generation, combined with ProteinMPNN\cite{RN2} for sequence design. 
While diffusion-based approaches are powerful, they often require specialized modifications 
for multi-objective design. In our example, the swarm framework was tasked with simultaneously 
achieving two objectives: introducing flexibility via turn-promoting residues 
and generating a repeating sequence pattern. Starting from a poly-valine sequence, 
the swarm designed a sequence (GGPPIGIGGIGGPGIIIGGGG) that met both criteria (mostly
loops and repeating GG pairs). This result demonstrates the swarm framework’s capability 
for multi-objective optimization.

These comparisons underscore the design capability, 
adaptability, and generalizability of our swarm framework for diverse protein design tasks.

\subsubsection*{Swarm Explores Novel Regions of Sequence Space}
To evaluate the novelty and biological plausibility of sequences generated by the swarm 
framework, we compared them to both naturally occurring proteins and 
sequences generated by established \textit{de novo} design methods.

We compiled a dataset of sequences consisting of: 
\begin{itemize}
    \item 5,000 natural protein sequences from the SCOPe database (v2.08)\cite{RN50}
    \item 200 sequences generated by ProteinMPNN\cite{RN2}
    from 100 backbones generated by RFdiffusion\cite{RN1} (two sequences per backbone) with unconditioned diffusion
    \item 640 sequences generated by our swarm framework across 10 distinct design objectives (64 sequences per objective)
\end{itemize}

\begin{figure}[htbp]
    \centering
    \includegraphics[width=\textwidth]{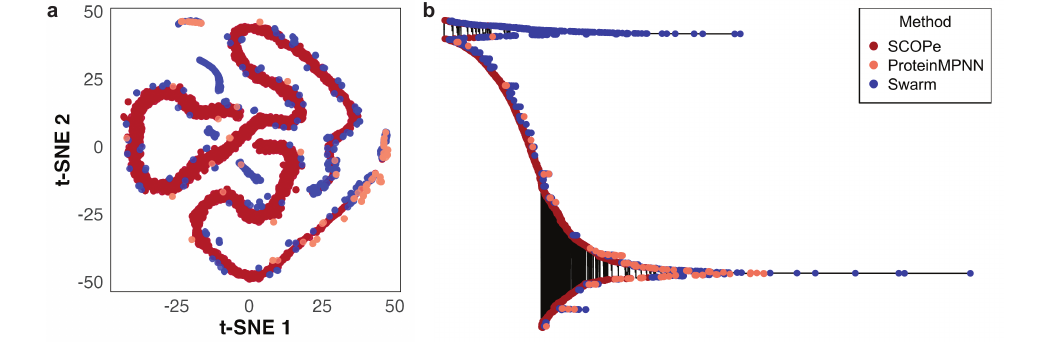}
    \caption{\textbf{a}. \textit{t}-SNE visualization 
    comparing swarm sequences (blue), SCOPe natural proteins (red), and 
    ProteinMPNN designs (wheat). \textbf{b}. 
    Tree using neighbor-joining on numerical feature vectors with the same 
    legend as in \textit{t}-SNE visualization.}
    \label{fig:sequence_space}
\end{figure}

Figure~\ref{fig:sequence_space}a visualizes how swarm framework, 
ProteinMPNN, and natural protein sequences from the SCOPe database distribute in a 
unified feature space, using t-distributed Stochastic Neighbor Embedding (\textit{t}-SNE) 
for dimensionality reduction. Each protein sequence was 
encoded as a feature vector encompassing amino acid composition, molecular weight, and aromaticity. The swarm framework samples sequence space broadly, generating sequences that are 
natural-like, others resembling \textit{de novo} designs from ProteinMPNN, and, 
notably, an additional distinct set that occupies regions unpopulated by either SCOPe or 
ProteinMPNN sequences. 

The tree in Figure~\ref{fig:sequence_space}b visualizes the
hierarchical relationships among sequences generated by the swarm framework, 
ProteinMPNN, and those derived from natural proteins in the SCOPe database. 
Construction of the tree is based on the same 22-dimensional numerical feature vectors. 
Sequence-to-sequence dissimilarities were computed,
and the neighbor-joining algorithm was employed to infer the tree topology. The tree reveals that the swarm framework produces a wide array of sequence 
types, including those near natural-like proteins and \textit{de novo} designs from ProteinMPNN, 
and others branching into novel regions of the sequence space. These 
results demonstrate that the swarm framework can generate both familiar and 
novel protein designs, extending beyond the reach of current natural and 
model-generated proteins.

\subsubsection*{Swarm Framework is Computationally Efficient}
A key advantage of the swarm framework is its computational efficiency, when 
compared to leading protein structure prediction and design models (Table~\ref{tab:computational_comparison_table}).

\begin{table}[htbp]
    \centering
    \begin{tabular}{@{}c c c@{}}
        \toprule
        \textbf{Model} & \textbf{Training Time} & \textbf{Inference Time per Prediction} \\
        \midrule
        AlphaFold\cite{RN48, RN49} & about 1400 TPU-days & minutes to hours \\
        ProtGPT2\cite{RN3} & 512 GPU-days & minutes \\
        ESM2\cite{RN54, RN53, RN55} & about 1800 GPU-days & 10 GPU-hours \\
        RFdiffusion\cite{RN1} & about 1800 GPU-days & minutes \\
        ProteinMPNN\cite{RN2} & about 10 GPU-days & minutes \\
        Swarm (this work) & no training required & a few GPU-hours \\
        \bottomrule
    \end{tabular}
    \caption{Comparative compute resource for protein structure prediction and design models, comparing training time and inference time per prediction.}
    \label{tab:computational_comparison_table}
\end{table}

Unlike methods such as AlphaFold (about 1400 TPU-days)\cite{RN48, RN49}, 
ProtGPT2 (512 GPU-days)\cite{RN3}, ESM2 (about 1800 GPU-days)\cite{RN54, RN53, RN55}, RFdiffusion (about 1800 GPU-days)\cite{RN1}, 
and ProteinMPNN\cite{RN2} which rely on heavy pretraining, 
the swarm framework requires no training. By eliminating the need for pretraining, 
the swarm framework lowers the computational overhead and democratizes 
access to \textit{de novo} protein design capabilities.

For inference, the swarm framework completes the full iterative design process 
within a few GPU-hours, with total runtime primarily determined by language model 
API response time. Compared to ESM2\cite{RN54, RN53, RN55} which may take approximately 10 GPU-hours per prediction, 
the swarm framework’s GPU-hour estimate is more cost-effective.
This combination of zero training cost and efficient inference renders the swarm framework an accessible and scalable approach for objective-directed protein design.

\section{Conclusion}
We introduced a decentralized, collaborative agent-based 
framework for protein sequence design, inspired by swarm intelligence. 
By assigning individual LLM agents to specific residue positions and allowing 
them to iteratively propose context-aware mutations, the swarm framework 
demonstrates versatility across diverse design objectives. 

Structurally, the swarm framework achieved precise designs of $\alpha$-helices, $\beta$-strands, and coils,
with high fidelity to residue conservation patterns (Figure~\ref{fig:best_sequences}). 
Alpha-helices and coils were validated by CD spectroscopy (Figure~\ref{fig:CD}). 
Functionally, the swarm framework engineered sequences to match the predefined vibrational frequency spectrum, 
satisfy metal-binding pockets, and form multi-domain structures (Figure~\ref{fig:more_design_objectives}). 
The iterative design process enables both convergence and exploration 
(Figure~\ref{fig:convergence},~\ref{fig:freq_convergence}).
To tune these design dynamics, different LLMs can be used to control the 
convergence and exploration behaviors (Figure~\ref{fig:model_comparison}).

Benchmarking against state-of-the-art protein engineering methods, 
including structure prediction models, autoregressive PLMs, and diffusion models, 
highlighted the swarm framework’s strengths in precise design control, adaptability on-the-fly, and
multi-objective optimization (Figure~\ref{fig:generalization_comparison}).
More importantly, the designed sequences explore biologically plausible and novel
regions of protein space (Figure~\ref{fig:sequence_space}). In addition, the swarm framework
requires no training and completes the design process within a few GPU-hours (Table~\ref{tab:computational_comparison_table}).

The swarm framework departs from traditional, monolithic deep-learning paradigms as
it employs a no-training generative approach where complex, multi-objective optimization 
emerges from the collaborative, local interactions of multiple specialized agents (Figure~\ref{fig:overview}). 
This swarm-of-agents principle is highly generalizable and holds considerable 
promise for other complex design domains, where locally-informed agents collaborate to achieve global objectives.

\section{Methods}
Codes and data are available at \href{https://github.com/lamm-mit/ProteinSwarm}{https://github.com/lamm-mit/ProteinSwarm}.
The swarm framework runs on a workstation equipped with a high-end CUDA-compatible GPU (Quadro RTX 5000, NVidia, Santa Clara, CA, USA).

\subsection{Swarm Framework}
The swarm framework consists of four core components: a multi-agent swarm system where each residue position is managed by an autonomous LLM agent, a memory and learning system that tracks global and local patterns, a structure evaluation unit with OmegaFold\cite{OmegaFold} for structure prediction and Rosetta\cite{RN34} for energy scoring, and a decision-making algorithm that accepts or rejects complete iterations based on structural and energetic criteria.

Each protein sequence is represented as a grid where each position $i$ in the sequence $S = (s_1, s_2, \ldots, s_n)$ is managed by an autonomous agent $A_i$. The agents operate on the amino acid sequence space $\mathcal{A}^{20}$, where each position can be occupied by any of the 20 standard amino acids. Each agent receives local context including linear sequence neighbors within a radius $r$: $\mathcal{N}_i = \{s_{i-r}, \ldots, s_{i-1}, s_{i+1}, \ldots, s_{i+r}\}$, spatial neighbors from the previous iteration's structure: $\mathcal{S}_i = \{(j, d_{ij}) : d_{ij} < \text{cutoff}\}$, solvent exposure information: $E_i = \text{exposure}(i, S, D)$, position-specific structural context from secondary structure analysis using DSSP software\cite{Joosten2011PDB}, and memory-based learning insights from previous iterations, where $D$ is the C$\alpha$ distance matrix from the previous iteration's folded structure.

The optimization proceeds through a cleanly separated four-phase loop:

\textbf{Phase 1: Agent Collection.} All agents simultaneously propose amino acid changes based on their local context and memory insights. Each agent $A_i$ receives input context $\mathcal{C}_i = (\mathcal{N}_i, \mathcal{S}_i, E_i, M_i)$, where $M_i$ represents memory-based learning insights, and outputs a proposed amino acid $a_i' \in \mathcal{A}^{20}$. The system collects all proposals $\mathbf{P} = (a_1', a_2', \ldots, a_n')$ without any structure computation during this phase.

\textbf{Phase 2: Apply Changes.} The proposed sequence $S' = (a_1', a_2', \ldots, a_n')$ is constructed from all agent proposals and the structure is predicted using OmegaFold:

\begin{equation}
    S' \xrightarrow{\text{OmegaFold}} PDB'
\end{equation}

The folding process used CUDA GPU acceleration with a subbatch size of 1 optimized for short sequences, outputting standard PDB coordinates.

\textbf{Phase 3: Structure Evaluation.} The folded structure is evaluated using multiple criteria including Rosetta energy\cite{RN34} scoring ($E_{\text{total}} = E_{\text{vdw}} + E_{\text{hbond}} + E_{\text{elec}} + \ldots$), secondary structure analysis using DSSP\cite{Joosten2011PDB}, and design objective-specific evaluation metrics.

\textbf{Phase 4: Decision and Memory Update.} The system decides whether to accept the proposed sequence based on:
\begin{equation}
    \text{Accept} = \begin{cases}
        \text{True} & \text{Objective Score}(S') > \text{Objective Score}(S) \\
        \text{True} & \text{if } E_{\text{total}}(S') < E_{\text{total}}(S) \text{ and } \text{Objective Score}(S') \approx \text{Objective Score}(S) \\
        \text{False} & \text{otherwise}
    \end{cases}
\end{equation}

If accepted, $S \leftarrow S'$ and the memory system is updated with the successful pattern. If rejected, the original sequence is retained and the memory system records the failed attempt.

The framework implements a memory system that enables agents to learn from both global and local patterns.

\textbf{Global Memory.} The framework tracks system-wide patterns including accepted and rejected sequences ($\mathcal{S}_{\text{accepted}}$, $\mathcal{S}_{\text{rejected}}$), successful mutation patterns ($\mathcal{P}_{\text{success}} = \{(p, c) : \text{success\_rate}(p) > \theta\}$), energy progression trends ($\mathcal{T}_{\text{energy}} = \{(i, E_i) : i \in \text{iterations}\}$), and structure score trends ($\mathcal{T}_{\text{structure}} = \{(i, S_i) : i \in \text{iterations}\}$).

\textbf{Local History.} Each agent maintains personal history tracking including personal action records ($\mathcal{A}_i = \{(a, \text{outcome}) : \text{position } i\}$), success rates for different amino acid substitutions, context-specific performance patterns, and neighboring position interaction effects.

\textbf{Memory Context Generation.} For each agent decision, the system provides:
\begin{equation}
    M_i = f(\mathcal{G}, \mathcal{L}_i, \mathcal{C}_i)
\end{equation}
where $\mathcal{G}$ is global memory, $\mathcal{L}_i$ is local history for position $i$, and $\mathcal{C}_i$ is current context.

Each agent $A_i$ receives a structured input including the current amino acid state ($s_i$), local sequence context ($\mathcal{N}_i$), spatial structural context ($\mathcal{S}_i$), solvent exposure ($E_i$), memory insights ($M_i$), design objective ($G$), and previous iteration outcomes. The agent outputs a structured proposal:
\begin{equation}
    \text{Proposal}_i = \{\text{reasoning}, \text{proposed\_value}\}
\end{equation}

where \text{proposed\_value} is constrained to a single amino acid code from $\mathcal{A}^{20}$.

\textbf{Energy Calculation:} Rosetta computes detailed energy terms including van der Waals interactions, hydrogen bonding, electrostatic interactions, and reference energies:
\begin{equation}
    E_{\text{total}} = \sum_{i} E_{\text{vdw}}(i) + \sum_{i,j} E_{\text{hbond}}(i,j) + \sum_{i,j} E_{\text{elec}}(i,j) + E_{\text{reference}}
\end{equation}

\textbf{Secondary Structure Analysis:} DSSP assigns secondary structure elements to each residue, classifying them as $\alpha$-helix (H), $\beta$-strand (E), or loop/coil (L):
\begin{equation}
    \text{SS}(i) = \begin{cases}
        H & \text{$\alpha$-helix} \\
        E & \text{$\beta$-strand} \\
        L & \text{loop/coil}
    \end{cases}
\end{equation}

\textbf{Objective-Specific Evaluation:} Custom metrics assess design objective achievement based on secondary structure composition, spatial arrangement, and energy terms:
\begin{equation}
    \text{ObjectiveScore} = f(\text{SS\_composition}, \text{spatial\_arrangement}, \text{energy\_terms})
\end{equation}






Each objective is evaluated using domain-specific metrics that combine structural analysis, energy scoring, and sequence composition analysis to provide comprehensive feedback to the swarm framework.

The system uses a recursion guard to prevent structure computation during agent processing, ensuring clean phase separation. Memory is persisted between iterations to enable learning, and all structural data is cached to minimize redundant computations.

\subsection{Experimental Validation}
Peptides were chemically synthesized and purchased from Genscript 
(Piscataway, NJ, U.S.A.) at 98\% purity followed by high-performance 
liquid chromatography. CD spectra were acquired using a peptide concentration 
of 1~mg/mL (SDEEDAAAQAKETESSES) dissolved in 0.1~M 
phosphate buffer (PB) and 0.1~mg/mL (KTEKTQQKTN) dissolved in 0.01~M PB respectively. 
The peptide concentration was chosen to stabilize the 
formation of secondary structures and enhance signal-to-noise, especially 
for detecting subtle helical content. CD measurements were performed in 
circular dichroism spectrophotometer (Jasco \#J-1500) in a standard 1-mm 
pathlength quartz cuvette (Jasco \#1103-0172) at room temperature, 
with wavelength scans from 260~nm to 190~nm at 0.5~nm interval. 
Each reported spectrum is the average of five (helix) and three (coil) scans, 
and baseline correction was performed by subtracting the 0.1~M or 0.01~M PB buffer 
spectrum. The spectra were analyzed on the BESTSEL online server to determine 
secondary structures\cite{RN62}.

\subsection{Sequence Space Analysis}

To evaluate the novelty and quality of swarm-generated sequences, we performed comprehensive sequence space analysis comparing swarm trajectories with ProteinMPNN-generated sequences.
We extracted unique sequences from ten swarm trajectory datasets:


\begin{itemize}
    \item form alpha helices using alanine, leucine, and glutamate in repeating patterns
    \item design one helix with one side hydrophobic and one side polar
    \item design beta strands by placing alternating hydrophobic and polar residues
    \item stabilize alpha helices using N-cap with serine, threonine and C-cap with glycine, proline, asparagine, aspartate
    \item form loose, extended coils using polar and charged residues to reduce compaction
    \item form a beta hairpin with two beta strands and one turn, using at least two aromatic residues in the strands and one proline in the turn
    \item promote compact local packing using hydrophobic residues and turn-promoting motifs
    \item create a helix-turn-helix motif with one helix followed by one turn and a second helix
    \item choose residues that mirror their left and right neighbors to promote local symmetry
    \item design protein with vibrational frequency spectrum matching target=[0.1, 0.15, 0.5, 0.6, 0.7, 0.8] for optimal dynamical properties
\end{itemize}

For comparison, we used ProteinMPNN-generated sequences as a 
baseline representing state-of-the-art deep learning protein design methods. All sequences were de-duplicated to ensure unique analysis.

For each sequence $S$, we computed a total of 22 numerical features: 20 amino acid composition features ($C(S) = [c_A, c_R, \ldots, c_Y]$ where $c_X$ is the frequency of amino acid $X$, representing the proportion of each of the 20 standard amino acids in the sequence), and 2 physicochemical properties (molecular weight, calculated as the average molecular weight per residue in Daltons, and aromaticity, representing the proportion of aromatic amino acids F, W, and Y). This feature set ensures fair comparison across all sequence types, as all sequences (SCOPe, SWARM, and ProteinMPNN) are treated equally based solely on sequence composition and physicochemical properties, without any structural bias that might penalize designed sequences lacking structural classifications. Data preprocessing involved robust feature selection to ensure numerical stability, including removal of near-zero variance features (threshold: $10^{-7}$) and elimination of highly correlated features (correlation threshold: 0.95).

\textit{t}-Distributed Stochastic Neighbor Embedding (\textit{t}-SNE) was performed to visualize sequence relationships in a 2D embedding space. The algorithm calculates pairwise similarities in the high-dimensional feature space and creates a 2D map that preserves local structure, where similar sequences cluster together and dissimilar sequences are pushed apart. We used optimized parameters for protein sequence analysis: perplexity of 30 to balance local and global structure preservation, maximum iterations of 1000 with early exaggeration, and random seed initialization for reproducibility. The perplexity parameter was set as $\min(30, \lfloor(N-1)/3\rfloor)$ where $N$ is the number of sequences. \textit{t}-SNE minimizes the Kullback–Leibler divergence between pairwise similarities in high- and low-dimensional spaces:

\begin{equation}
    \text{KL}(P \parallel Q) = \sum_{i \neq j} p_{ij} \log \frac{p_{ij}}{q_{ij}}
\end{equation}

where $p_{ij}$ denotes the similarity between data points $i$ and $j$ in the high-dimensional space, and $q_{ij}$ denotes their similarity in the low-dimensional embedding. The resulting 2D coordinates (\textit{t}-SNE1, \textit{t}-SNE2) for each sequence enable visualization where distance in 2D approximates sequence similarity, with clusters representing groups of sequences sharing similar properties and separation indicating different regions occupied by different methods (SCOPe, SWARM, ProteinMPNN).

Trees were constructed using the neighbor-joining (NJ) algorithm on Euclidean distance matrices computed from the feature vectors. The distance between sequences $i$ and $j$ was calculated as $d_{ij} = \sqrt{\sum_{k} (x_{ik} - x_{jk})^2}$ for feature vectors $x_i$ and $x_j$, where lower distance indicates more similar sequences. The NJ algorithm iteratively groups the nearest neighbors (most similar sequences) based on the minimum evolution principle, 
creating an unrooted tree that minimizes the total branch length while maintaining additivity of distances. For computational efficiency, datasets exceeding a size threshold were subsampled to 1000 sequences. The resulting tree was rooted at the first sequence (arbitrary rooting) to convert it to a rooted tree for visualization. It is important to note that this tree is based on feature similarity rather than evolutionary history, showing functional and structural relationships between sequences rather than true phylogenetic relationships.

\section*{Acknowledgments}
\textbf{Compute:} We acknowledge support from the MIT Office of Research Computing and Data for providing computational resources. 
\textbf{Funding:} We acknowledge support by the Bernard E. Proctor Memorial Fund and 2025 Mathworks Fellowship. 
\textbf{Author contributions:} M.J.B. and F.Y.W. conceived the idea. 
F.Y.W. and M.J.B. developed the framework, ran iterations, and analyzed the results. D.S.L. conducted CD spectroscopy experiments.
M.J.B. and D.L.K. supervised the project. F.Y.W. and M.J.B. wrote the manuscript. 
\textbf{Competing interests:} The authors declare that they have no competing interests. 
\textbf{Data and materials availability:} All data needed to evaluate the conclusions in the paper are 
present in the paper and the Supplementary Materials. 
Additional data related to this paper may be requested from the corresponding author.

\bibliographystyle{naturemag}

\bibliography{references}

\newpage
\section*{\Large Supplementary Materials}

\begin{center}
  \vspace{2in}
    \LARGE {Swarms of Large Language Model Agents for Protein Sequence Design with Experimental Validation}
\end{center}
\vspace{0.1in}
\begin{center}
    Fiona Y. Wang \quad
    Di Sheng Lee \quad
    David L. Kaplan \quad
    Markus J. Buehler\textsuperscript{3,\#}\\
    \vspace{0.1in}
    Corresponding author:
    \textsuperscript{\#}
    \texttt{mbuehler@mit.edu}
\end{center}
\vspace{1in}
\textbf{The PDF file includes:} 
\qquad Figures S1 to S5
\begin{figure}[htbp]
  \centering
  \includegraphics[width=0.85\textwidth]{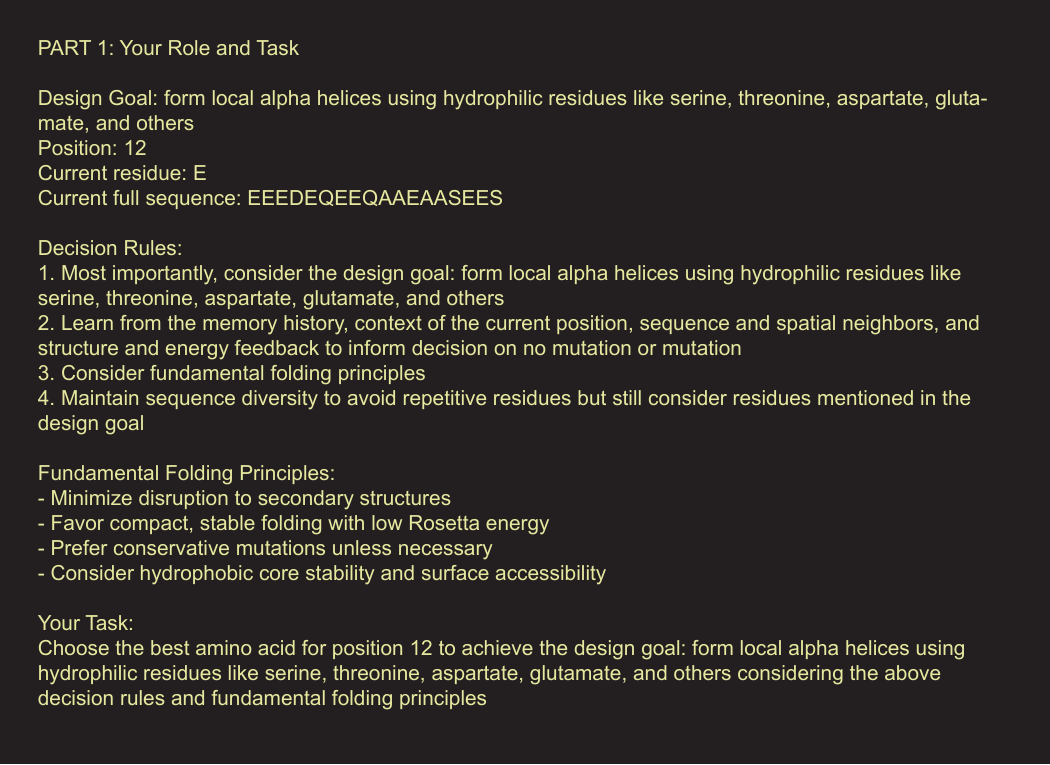}
  \caption*{Figure S1. Example input prompt (Part 1: Role and Task) given to one LLM agent.}
  \label{fig:figS1}
\end{figure}

\begin{figure}[htbp]
  \centering
  \includegraphics[width=0.85\textwidth]{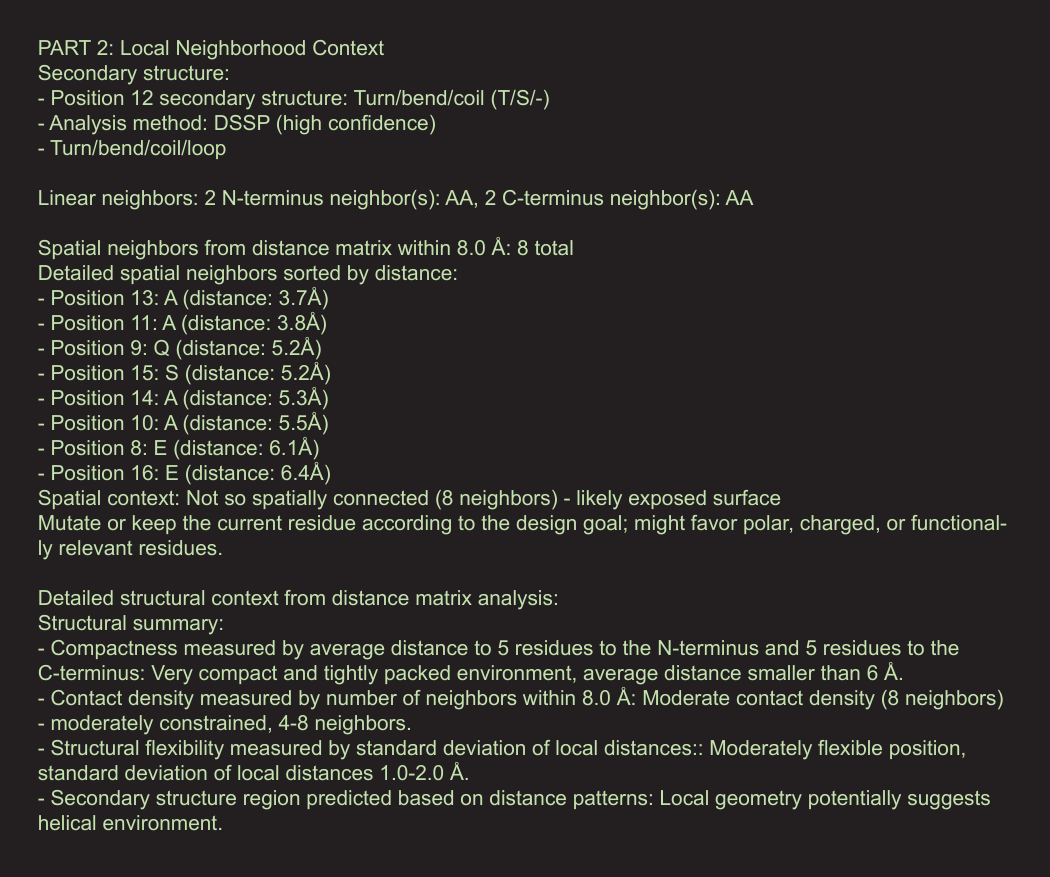}
  \caption*{{Figure S2.} Example input prompt (Part 2: Local Neighborhood Context) given to one LLM agent.}
  \label{fig:figS2}
\end{figure}

\begin{figure}[htbp]
  \centering
  \includegraphics[width=0.85\textwidth]{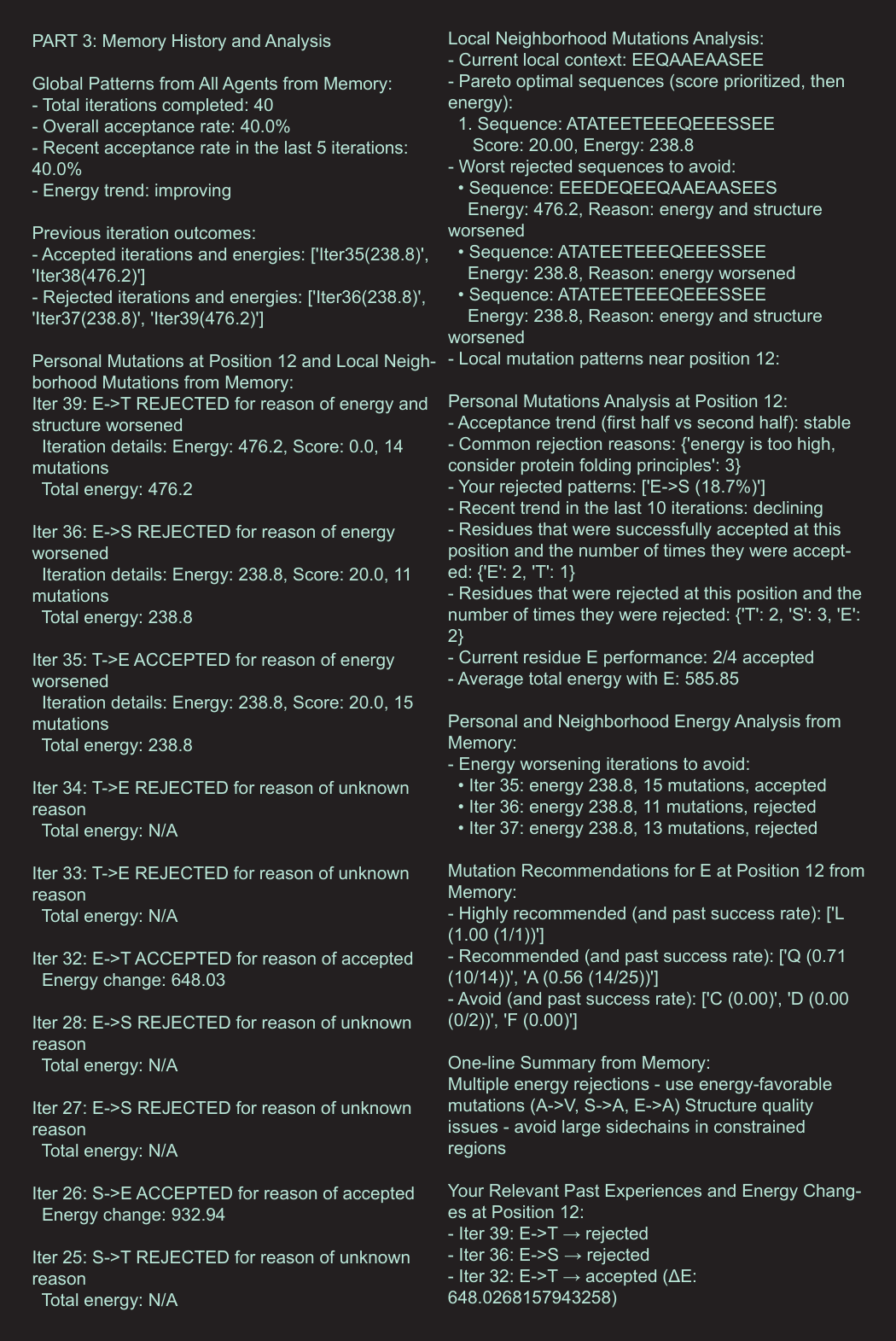}
  \caption*{{Figure S3.} Example input prompt (Part 3: Memory History and Analysis) given to one LLM agent.}
  \label{fig:figS3}
\end{figure}

\begin{figure}[htbp]
  \centering
  \includegraphics[width=0.85\textwidth]{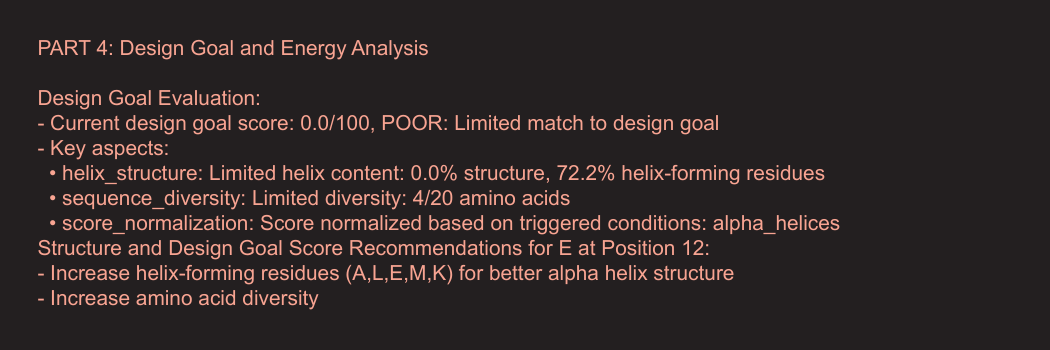}
  \caption*{{Figure S4.} Example output (Part 4: Design Goal and Energy Analysis) given to one LLM agent.}
  \label{fig:figS4}
\end{figure}

\begin{figure}[htbp]
  \centering
  \includegraphics[width=0.85\textwidth]{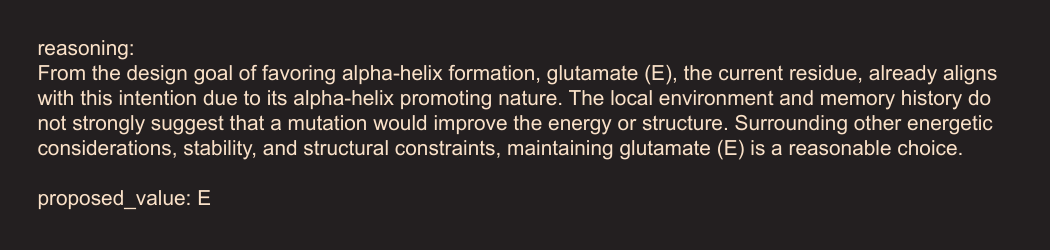}
  \caption*{{Figure S5.} Example output with reasoning and proposed mutation from one LLM agent.}
  \label{fig:figS5}
\end{figure}

\end{document}